\documentclass[letterpaper, 10 pt, conference]{ieeeconf}  

\IEEEoverridecommandlockouts                              

\usepackage{graphics} 
\usepackage{epsfig} 
\usepackage{times} 
\usepackage{amsmath} 
\usepackage{amssymb}  
\usepackage{hyperref}
\usepackage{bibentry}
\usepackage{microtype}
\usepackage{balance}
\hypersetup{hidelinks}
\usepackage{placeins}
\usepackage{xcolor}

\title{\LARGE \bf
iRonCub 3: The Jet-Powered Flying Humanoid Robot 
}

\author{Davide Gorbani$^{* 1,2}$, Hosameldin Awadalla Omer Mohamed$^{* 1}$, Giuseppe L'Erario$^{1}$, Gabriele Nava$^{1}$,\\ Punith Reddy Vanteddu$^{1,2}$, Shabarish Purushothaman Pillai$^{1}$, Antonello Paolino$^{1,3}$,\\ Fabio Bergonti$^{1}$, Saverio Taliani$^{1}$, Alessandro Croci$^{1}$, Nicholas James Tremaroli$^{1}$,\\ Silvio Traversaro$^{1}$, Bruno Vittorio Trombetta$^{1}$ and  Daniele Pucci$^{1,2}$
\thanks{$^{*}$These authors contributed equally to this work.}
\thanks{$^{1}$Artificial and Mechanical Intelligence, Istituto Italiano di Tecnologia, Genoa, Italy {\tt\small firstname.surname@iit.it}}%
\thanks{$^{2}$School of Computer Science, University of Manchester, Manchester, UK}%
\thanks{$^{3}$Department of Industrial Engineering, University of Naples Federico II, Naples, Italy}}

\begin{document}


\newcommand{\brRound}[1]{\left( #1 \right)}
\newcommand{\brSquare}[1]{\left[ #1 \right]}
\newcommand{\brCurly}[1]{\{ #1 \}}

\newcommand{\bm}[1]{\begin{bmatrix} #1 \end{bmatrix}}

\newcommand{\mx}[1]{\boldsymbol{#1}}

\newcommand{\vect}[1]{\boldsymbol{#1}}

\newcommand{\R}[1]{\mathbb{R}^{#1}}
\newcommand{\RR}[2]{\mathbb{R}^{#1 \times #2}}
\newcommand{\SO}[1]{\text{SO} \brRound{#1}}

\newcommand{\eye}[1]{{\mx{I}}_{#1}}

\newcommand{\zero}[1]{\boldsymbol{0}_{#1}}
\newcommand{\zeroX}[2]{\boldsymbol{0}_{#1 \times #2}}


\newcommand{\tr}{\top}
\newcommand{\trInv}{{-\top}}

\newcommand{\skw}[1]{\mathcal{S}{\brRound{#1}} }

\newcommand{\barSkw}[1]{\bar{\mathcal{S}}{\brRound{#1}} }

\newcommand{\rank}[1]{\text{rank} \brRound{#1} }


\newcommand{\nullSpace}[1]{\text{null} \brRound{#1} }

\newcommand{\nullSpaceMatrix}[1][]{\mathcal{Z}_{#1} }




\newcommand{\numberSphericalJoints}{N_{\text{sj}}}

\newcommand{\rotm}[2]{R^{#1}_{#2}}
\newcommand{\rotmW}[1]{\rotm{\frameW}{#1}}


\newcommand{\tform}[2]{T^{#1}_{#2}}

\newcommand{\quat}[2]{\mathcal{Q}^{#1}_{#2}}
\newcommand{\quatW}[1]{\quat{\frameW}{#1}}

\newcommand{\dquat}[2]{\dot{\mathcal{Q}}^{#1}_{#2}}
\newcommand{\dquatW}[1]{\dquat{\frameW}{#1}}

\newcommand{\versor}[3]{\boldsymbol{\hat{#1}}^{#2}_{#3}}
\newcommand{\versorW}[2]{\boldsymbol{\hat{#1}}^{\frameW}_{#2}}
\newcommand{\dversor}[3]{\boldsymbol{\dot{\hat{#1}}}^{#2}_{#3}}
\newcommand{\dversorW}[2]{\boldsymbol{\dot{\hat{#1}}}^{\frameW}_{#2}}

\newcommand{\pos}[3]{\boldsymbol{p}^{#1}_{#2,#3}}
\newcommand{\posW}[1]{\boldsymbol{p}_{#1}}

\newcommand{\linVel}[3]{\boldsymbol{v}^{#1}_{#2,#3}}
\newcommand{\linVelW}[1]{\boldsymbol{v}_{#1}}

\newcommand{\angVel}[3]{\boldsymbol{\omega}^{#1}_{#2,#3}}
\newcommand{\angVelW}[1]{\boldsymbol{\omega}_{#1}}

\newcommand{\node}[2]{\left( #1,#2 \right)}
\newcommand{\nodeij}{\node{i}{j}}
\newcommand{\nodeipj}{\node{i+1}{j}}
\newcommand{\nodeijp}{\node{i}{j+1}}
\newcommand{\nodeimj}{\node{i-1}{j}}
\newcommand{\nodeijm}{\node{i}{j-1}}

\newcommand{\J}[1]{J_{#1}}
\newcommand{\dJ}[1]{\dot{J}_{#1}}
\newcommand{\Jc}{\J{c}}
\newcommand{\dJc}{\dJ{c}}

\newcommand{\convm}[2]{\mathcal{M}^{#1}_{#2}}

\newcommand{\selm}[1]{S_{#1}}

\newcommand{\permm}[1]{P_{#1}}

\newcommand{\nodesAbsVel}[1][]{\boldsymbol{\chi}_{#1}}

\newcommand{\nodesAbsOmega}{\boldsymbol{\omega}}

\newcommand{\nodesAbsLinVel}{\boldsymbol{v}}

\newcommand{\nodesRelOmega}[1][]{\boldsymbol{\omega}^{R}_{r_{#1}}}

\newcommand{\nodesRelOmegaAct}{\nodesRelOmega[act]}

\newcommand{\nodesRelOmegaInertial}{\boldsymbol{\omega}^{0}_r}

\newcommand{\vb}{\boldsymbol{V_b}}

\newcommand{\VbAndNodesAbsOmega}{\boldsymbol{\nu_a}}

\newcommand{\VbAndNodesRelOmega}[1][]{\boldsymbol{\nu}_{#1}}

\newcommand{\VbAndNodesRelOmegaInertial}{\boldsymbol{\nu^0}}

\newcommand{\VbAndNodesRelOmegaEcon}{\boldsymbol{\nu_e}}

\newcommand{\distJointI}{\boldsymbol{l_{1}}}
\newcommand{\distJointJ}{\boldsymbol{l_{2}}}
\newcommand{\distNodes}{2 l}

\newcommand{\stateP}{\boldsymbol{x}}
\newcommand{\stateV}{\boldsymbol{\dot{x}}}

\newcommand{\momentum}[1]{{}_{\mathcal{#1}}\boldsymbol{L}}

\newcommand{\dmomentum}[1]{{}_{\mathcal{#1}}\dot{\boldsymbol{L}}}

\newcommand{\centmomentum}{{}_{\frameCent}\boldsymbol{L}}

\newcommand{\dcentmomentum}{{}_{\frameCent}\dot{\boldsymbol{L}}}

\newcommand{\jetsT}{\boldsymbol{T}}

\newcommand{\jetsdT}{\dot{\boldsymbol{T}}}

\newcommand{\frameW}{\mathcal{J}}

\newcommand{\frameCoM}{\mathcal{G}}

\newcommand{\frameB}{\mathcal{B}}

\newcommand{\frameLF}{\mathcal{LF}}

\newcommand{\frameRF}{\mathcal{RF}}

\newcommand{\frameLL}{\mathcal{LL}}

\newcommand{\frameRL}{\mathcal{RL}}

\newcommand{\frameCent}{\mathcal{\frameCoM[\frameW]}}

\newcommand{\wrenchX}[2]{{}_{\mathcal{#1}}^{}\mx{X}^{\mathcal{#2}} { }}

\newcommand{\velX}[2]{{}_{}^{\mathcal{#1}}\mx{X}_{\mathcal{#2}}}

\newcommand{\wrench}[1]{{}_{\mathcal{#1}}^{}\vect{f}}

\newcommand{\force}[1]{{}_{\mathcal{#1}}^{}\vect{f}}

\newcommand{\torque}[1]{{}_{\mathcal{#1}}^{}\vect{\tau}}

\newcommand{\dwrench}[1]{{}_{\mathcal{#1}}\dot{\vect{f}}}

\newcommand{\ddwrench}[1]{{}_{\mathcal{#1}}\ddot{\vect{f}}}

\newcommand{\homomx}[2]{{}_{}^{\mathcal{#1}}\mx{H}_{\mathcal{#2}} { }}

\newcommand{\rotmx}[2]{{}_{}^{\mathcal{#1}}\mx{R}_{\mathcal{#2}} { }}

\newcommand{\dotrotmx}[2]{{}_{}^{\mathcal{#1}}\dot{\mx{R}}_{\mathcal{#2}} { }}

\newcommand{\transrotmx}[2]{{}_{}^{\mathcal{#1}}\mx{R}^T_{\mathcal{#2}} { }}

\newcommand{\posv}[2]{{}_{}^{\mathcal{#1}}\vect{o}_{\mathcal{#2}} { }}

\newcommand{\dotposv}[2]{{}_{}^{\mathcal{#1}}\dot{\vect{o}}_{\mathcal{#2}} { }}

\newcommand{\ddotposv}[2]{{}_{}^{\mathcal{#1}}\ddot{\vect{o}}_{\mathcal{#2}} { }}

\newcommand{\velv}[2]{{}_{}^{\mathcal{#1}}\text{\textbf{v}}_{\mathcal{#2}} { }}

\newcommand{\dotvelv}[2]{{}_{}^{\mathcal{#1}}\dot{\text{\textbf{v}}}_{\mathcal{#2}} { }}

\newcommand{\jomosubv}[2]{{}_{}^{\mathcal{#1}}\vect{s}_{\mathcal{#2}} { }}

\newcommand{\linvelv}[2]{{}_{}^{\mathcal{#1}}\vect{\nu}_{\mathcal{#2}} { }}

\newcommand{\angvelv}[2]{{}_{}^{\mathcal{#1}}\vect{\omega}_{\mathcal{#2}} { }}

\newcommand{\dotangvelv}[2]{{}_{}^{\mathcal{#1}}\dot{\vect{\omega}}_{\mathcal{#2}} { }}

\maketitle
\thispagestyle{empty}
\pagestyle{empty}


\begin{abstract}

This article presents iRonCub 3, a jet-powered humanoid robot, and its first flight experiments. Unlike traditional aerial vehicles, iRonCub 3 aims to achieve flight using a full-body humanoid form, which poses unique challenges in control, estimation, and system integration. We highlight the robot’s current mechanical and software architecture, including its propulsion system, control framework, and experimental infrastructure. The control and estimation framework is first validated in simulation by performing a takeoff and tracking a reference trajectory. Then, we demonstrate, for the first time, a liftoff of a jet-powered humanoid robot — an initial but significant step toward aerial humanoid mobility. Also, we detail how the experimental area around a jet-powered humanoid robot should be designed in order to deal with a level of complexity that is substantially superior than indoor humanoid robot experiments.

\end{abstract}
\section{INTRODUCTION}

Humanoid robots are increasingly employed in research and industry due to their ability to operate in environments designed for humans. Traditionally, their capabilities have been confined to terrestrial locomotion and manipulation. However, the integration of aerial mobility into humanoid platforms presents a new and promising research frontier.

Several works have explored the integration of aerial mobility with manipulation and terrestrial locomotion. These efforts typically focus on equipping aerial robots with additional capabilities, such as robotic arms or wheeled/legged appendages, to enhance their versatility in interacting with complex environments~\cite{ruggiero2018aerial, LeoCaltechPaper, Jet-HR1}. A common example is the class of aerial manipulators, which consist of multirotor platforms outfitted with robotic arms or grippers for physical interaction~\cite{heredia2014aerialmanip, ruggiero2018aerial}. However, aerial manipulators represent just one subset of multibody flying robots.
Other notable examples include: hexapod-quadrotor hybrid systems~\cite{pitonyak2017hexapod}; platforms that combine aerial and terrestrial locomotion, such as hybrid quadrotors with wheels or legs~\cite{kalantari2013hytaq}; bio-inspired insect robots capable of both flight and ground movement~\cite{bozkurt2009insectbio}; and reconfigurable structures that can morph to adapt to different locomotion modes~\cite{daler2015multimodal, daler2013morphology}. Furthermore, recent research has investigated humanoid robots enhanced with thrusters to support bipedal motion and short bursts of flight~\cite{LeoCaltechPaper, Jet-HR1}.

\begin{figure}[t]
    \centering
    \includegraphics[width=1.0\linewidth]{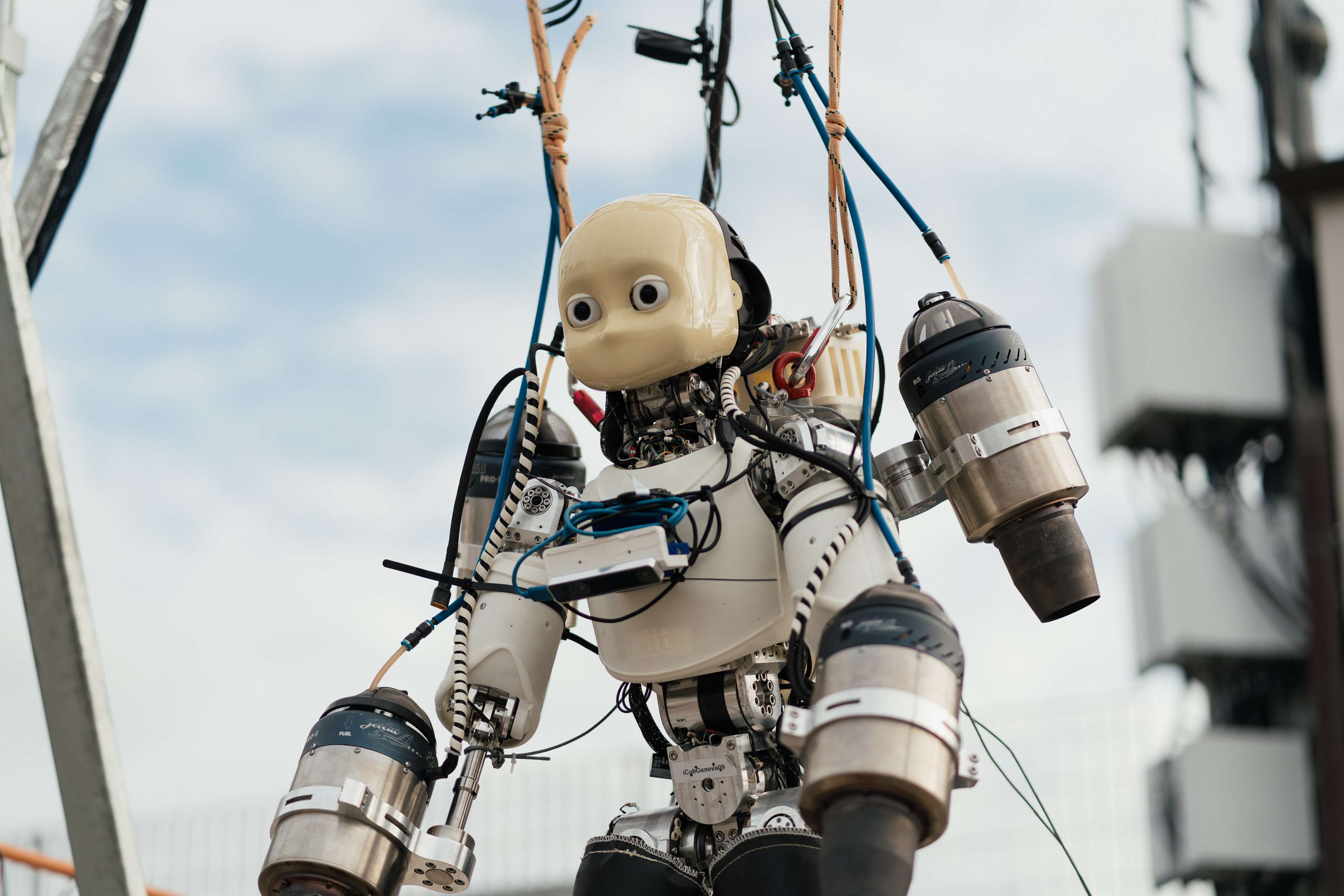}
    \caption{iRonCub-Mk3 in the experimental area.}
    \label{fig:ironcub}
\end{figure}

A more ambitious direction is represented by the iRonCub project~\cite{pucci2017momentum, nava2018position}, which seeks to unify aerial, terrestrial, and manipulation capabilities within a single humanoid platform. The goal is to make the iCub humanoid robot~\cite{dafarra2024icub3} fly with the addition of four jet turbines, two mounted on the arms and two on the jetpack. The potential applications of aerial humanoid robots extend across multiple high-impact domains where ground-based or conventional aerial systems face limitations. In search and rescue missions, for instance, flying humanoid robots can traverse collapsed buildings, steep terrain, or flooded zones, combining flight with the ability to interact physically with the environment, such as opening doors or lifting debris. In infrastructure inspection and maintenance, a flying humanoid could reach hard-to-access locations like bridges, high-rise structures, or industrial plants, and perform tasks requiring dexterity, such as turning valves or manipulating tools.

Another promising area is disaster response in hazardous environments, such as nuclear facilities or chemical plants, where robots must operate in complex spaces designed for humans while avoiding harmful exposure. Additionally, applications in telepresence and human-avatar systems could benefit from a robot that can navigate complex 3D environments while mimicking human motion, useful for virtual interaction, inspection, or even remote training scenarios.\looseness=-1

By leveraging the human-like morphology, aerial humanoid robots offer the promise of combining the spatial access and reach of flying robots with the intuitive physical interaction capabilities of humanoids, potentially opening new frontiers in robotics deployment.

In this article, we outline the current status of the iRonCub project, covering the latest design version, referred to as iRonCub-Mk3, a jet-powered humanoid robot developed as a research platform to investigate aerial humanoid locomotion. The iRonCub-Mk3 represents an advancement over the earlier model, iRonCub-Mk1, which was presented in the article~\cite{Mohamed2022thrust}.

This article is showcasing:

\begin{itemize}
    \item The design and integration of a custom jetpack and modified forearms into the iCub3 platform;
    \item A software architecture for estimation and control that leverages model-based approaches;
    \item An experimental demonstration of vertical liftoff — the first step toward fully autonomous humanoid flight.
\end{itemize}

The remainder of this article is organised as follows: Section \ref{sec:methods} describes the system architecture and methods; Section \ref{sec:expArea} outlines the experimental infrastructure; Section \ref{sec:results} presents simulation and hardware results; and Section \ref{sec:conclusion} concludes with a discussion of limitations and future work.

\section{METHODS}
\label{sec:methods}

\subsection{Mechanical design}

The iRonCub-Mk3 platform is based on the iCub3 humanoid robot \cite{dafarra2024icub3} and is equipped with four JetCat P250 Pro turbines\footnote{www.jetcat.de/en/products/produkte/jetcat/kategorien/professional/P250-PRO-S}, two mounted on the forearms and two integrated into a custom-designed jetpack. Fig. 2 shows the CAD model of the iRonCub-Mk3 configuration, highlighting the robot’s design, including the forearm turbine placement and jetpack integration.

\begin{figure}
    \centering
    \includegraphics[trim=450 200 700 200, clip,width=0.65\linewidth]{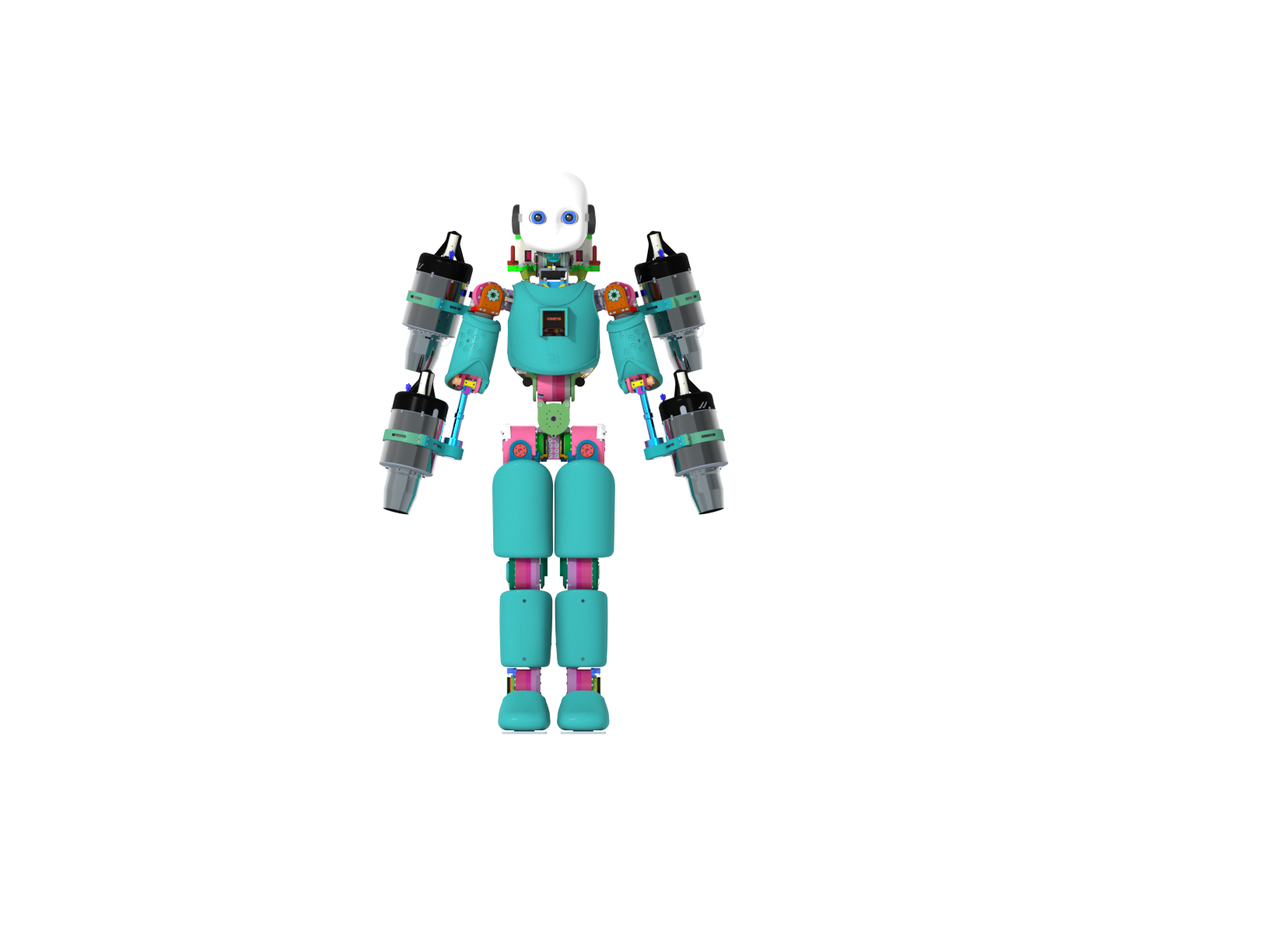}
    \caption{CAD rendering of iRonCub-Mk3.}
    \label{fig:ironcub-cad}
\end{figure}

\subsubsection{Jetpack design}

Building upon the iCub3 design \cite{dafarra2024icub3}, modifications were made to securely mount the jetpack. Specifically, load-bearing mounting brackets were custom-manufactured and installed on both the left and right sides of the central block housing the shoulder pitch motors and torso yaw. These brackets provide additional fixture points for securely attaching the jetpack. The integration of the jetpack with these components is shown in Fig. \ref{fig:icub3-jetpack}. The jetpack design was driven by the following considerations:

\begin{figure*}
    \centering
    \includegraphics[trim=0 15 0 0, clip,width=1.0\linewidth]{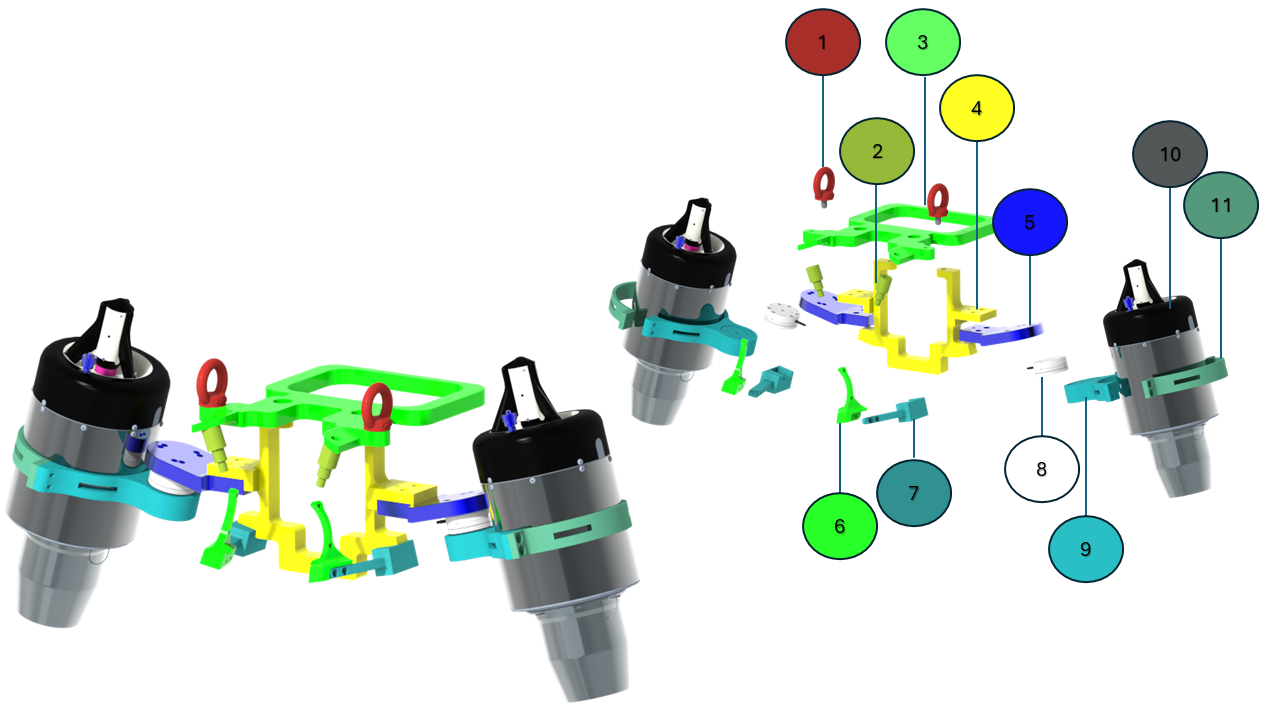}
    \caption{Exploded view (right) and assembled view (left) of the jetpack. The components labeled numerically and color coded in the figure: 1. eyelet screw, 2. load-bearing fixture, 3. top plate 4. vertical frame, 5. Force-torque sensor bracket, 6. mounting bracket installed on the robot to provide additional fixture, 7. Connection between frame and mounted bracket, 8. Force-torque sensor, 9. outer turbine clamp, 10. Jetcat P250 Pro turbine, and 11. inner turbine clamp. }
    \label{fig:icub3-jetpack}
\end{figure*}

\begin{figure*}
    \centering
    \includegraphics[trim=0 100 0 40, clip,width=1.0\linewidth]{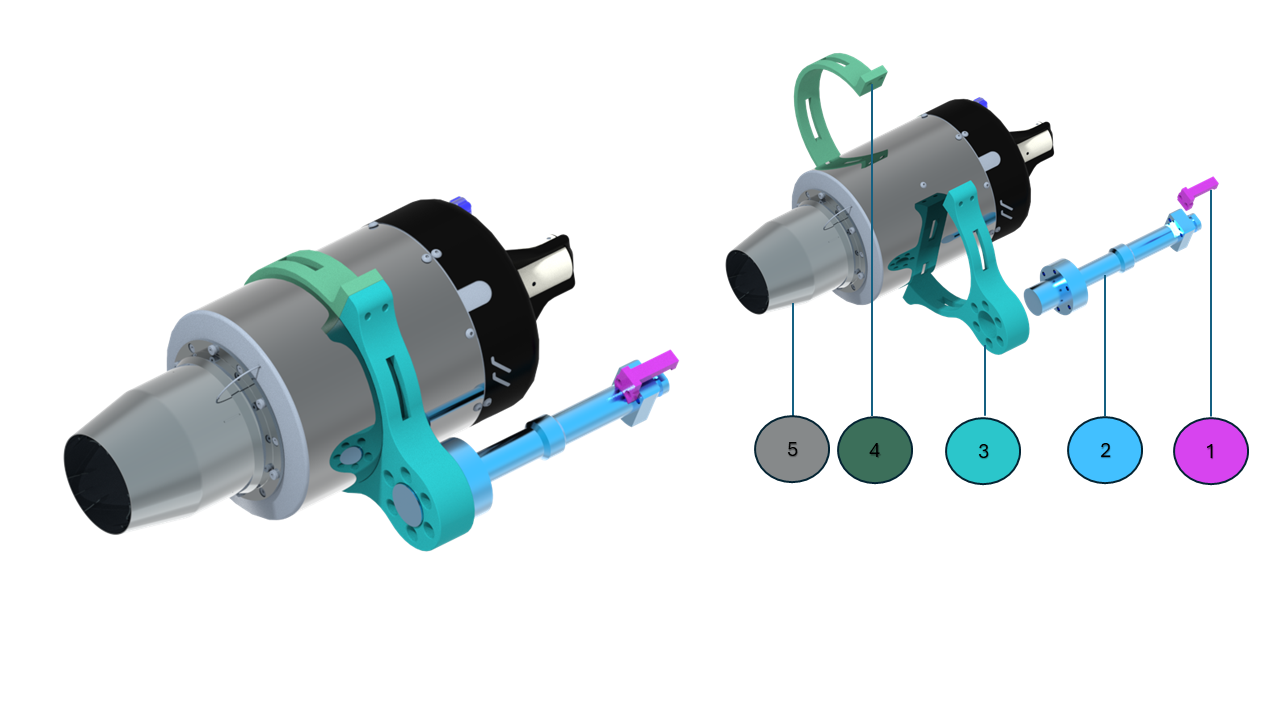}
    \caption{CAD rendering of the forearms of the iRonCub-Mk3 robot. In this version, the forearms of the iCub robot are removed to facilitate flight tests. On the left is the assembled view, and on the right is the exploded view. The components labeled numerically and color coded in the figure include: 1. forearm lock, 2. forearm shaft, 3. turbine inner clamp, 4. Jetcat P250 Pro turbine, and 5. turbine outer clamp.}
    \label{fig:ironcub-forearm}
\end{figure*}

\begin{itemize}
    \item Ease of assembly, disassembly.
    \item Minimal volume.
    \item Structural integrity to handle the stress forces exerted by the jet engines.
    \item Minimal modification to the original iCub3 design.
\end{itemize}

To enhance stability during flight and improve control over angular momentum, slight tilt angles were incorporated into the attached jet engines. This tilt also helps mitigate the impact of exhaust heat on the robot’s lower body. 

\begin{figure*}
    \centering
    \includegraphics[trim=20 110 50 180, clip,width=0.9\linewidth]{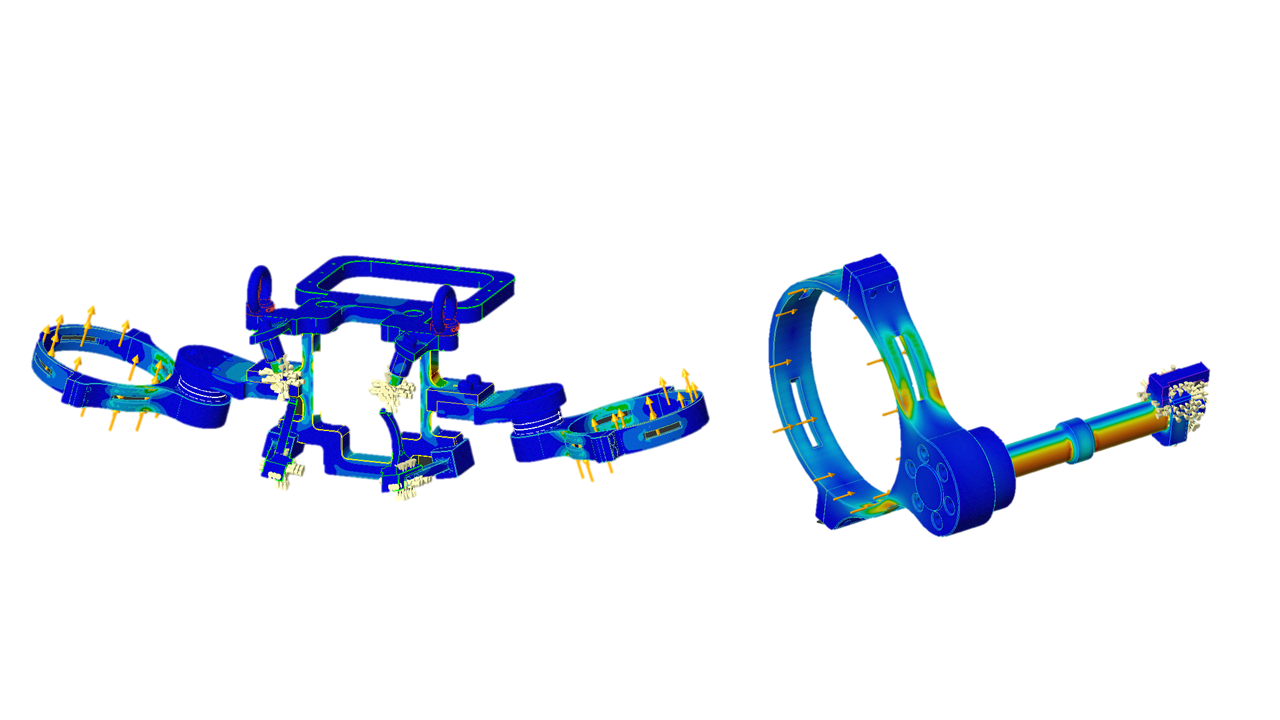}
    \caption{Illustration of the FEM analysis on the jetpack (left) and forearm (right). Boundary conditions were fixed at the screw points on the robot, and load conditions were set to $750 N$ on the clamps of the jet engines, axially loaded, for both components.
}
    \label{fig:fem-jetpack}
\end{figure*}

\subsubsection{Forearms design}

The iRonCub-Mk3 prototype was initially configured without forearms to streamline the development of flight control systems. In this version, jets are mounted on the arms below the elbows, providing thrust for stabilisation and manoeuvrability. The hands were removed from this prototype to focus exclusively on testing the flight capabilities, rather than manipulation at this stage. However, the forearms remain compatible with the system and can be integrated in later versions as needed. Fig. \ref{fig:ironcub-forearm} shows the CAD rendering of the forearms of the iRonCub-Mk3.

Additionally, a Finite Element Method (FEM) analysis was conducted on both the jetpack and the forearm to ensure their structures could withstand the forces from the jet engines. Boundary conditions were fixed at the assembly fixture locations on the robot, and axial loads of $750 N$, which is three times the peak thrust of each turbine, were applied to the brackets of the jet engines. Fig.~\ref{fig:fem-jetpack} illustrates the results of this FEM analysis.

\subsection{Software architecture}

\begin{figure*}
    \centering
    \includegraphics[width=0.95\linewidth]{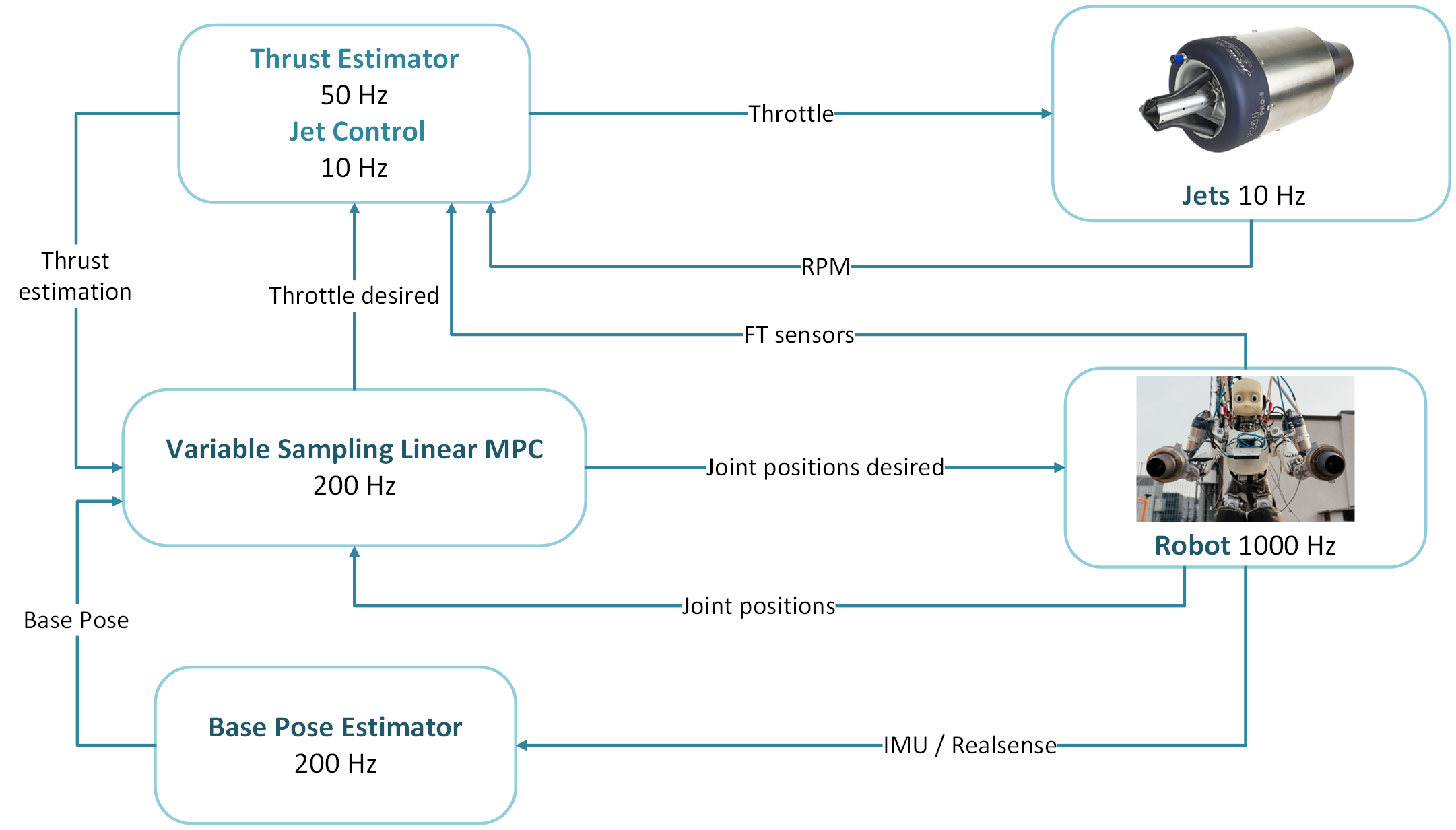}
    \caption{Block diagram showing the interaction among the main software components. The \textbf{Thrust Estimator} and \textbf{Jet Control}, described in \autoref{subsec:thrustEstim}, communicate with the jet engines and estimate the generated thrust using sensor data from both the robot and the turbines. The \textbf{Base Pose Estimator}, outlined in \autoref{subsec:basePoseEstim}, calculates the robot's base pose using information from onboard sensors. The \textbf{Flight Controller}, explained in \autoref{subsec:flightController}, implementing a \textbf{Variable Sampling Linear Model Predictive Controller (MPC)}, generates the required throttle commands for the jet turbines along with the target joint positions.}
    \label{fig:swArc}
\end{figure*}

The high-level software architecture is based on three components:

\begin{itemize}
    \item Base Pose Estimator
    \item Thrust Estimator and Jet Control
    \item Flight Controller
\end{itemize}

Fig.~\ref{fig:swArc} illustrates the interaction between the various software components. The \textbf{Thrust Estimator} and \textbf{Jet Control}, described in \autoref{subsec:thrustEstim}, interface with the jet engines and estimate the generated thrust using measurements from both the robot and the jet turbines. The \textbf{Base Pose Estimator}, discussed in \autoref{subsec:basePoseEstim}, determines the robot's base pose based on data from onboard sensors. The \textbf{Flight Controller}, detailed in \autoref{subsec:flightController}, computes the desired throttle commands for the jet turbines as well as the target joint positions.

Throttle commands are transmitted to the jet turbines’ low-level controller, which operates at $10~Hz$, while the desired joint positions are sent to the joint-level low-level controller, running at $1000~Hz$.

All the software components run on the robot's onboard computer, with communication between them handled by the YARP middleware~\cite{fitzpatrick2014middle}.

\subsection{Thrust estimation and Jet Control}
\label{subsec:thrustEstim}

To estimate the thrust produced by the jet engines, we implemented an Unscented Kalman Filter (UKF) that combines measurements from force-torque sensors mounted on both the robot’s arms and jetpack, with a second-order nonlinear dynamic model that predicts thrust output based on throttle input values. The force-torque sensors were calibrated in situ following the procedure outlined in \cite{mohamed2023NonlinearFTCalib}, enhancing their accuracy across a broader range of force and torque measurements.

\begin{figure}
    \centering
    \includegraphics[width=0.95\linewidth]{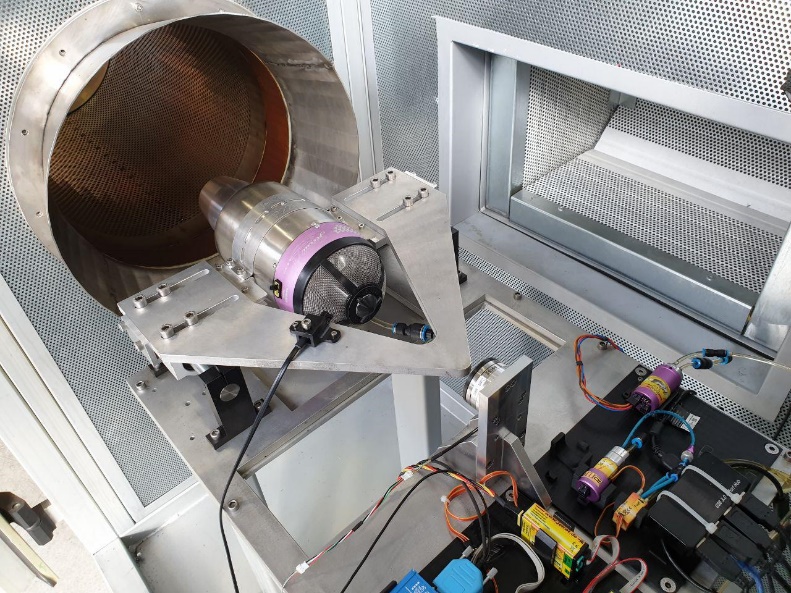}
    \caption{Custom-designed testbench used for conducting experiments on the jet engines to develop dynamic models of thrust generation.}
    \label{fig:test-bench}
\end{figure}

To develop the dynamic model for thrust prediction, a dedicated test bench was constructed, as shown in Fig.~\ref{fig:test-bench}. This setup enabled controlled experiments on the jet engines, generating datasets used to identify models that relate input commands to the resulting thrust levels. Further details of these studies can be found in \cite{lerario2020jets, momin2022turbines}.

\begin{figure}
    \centering
    \includegraphics[width=0.95\linewidth]{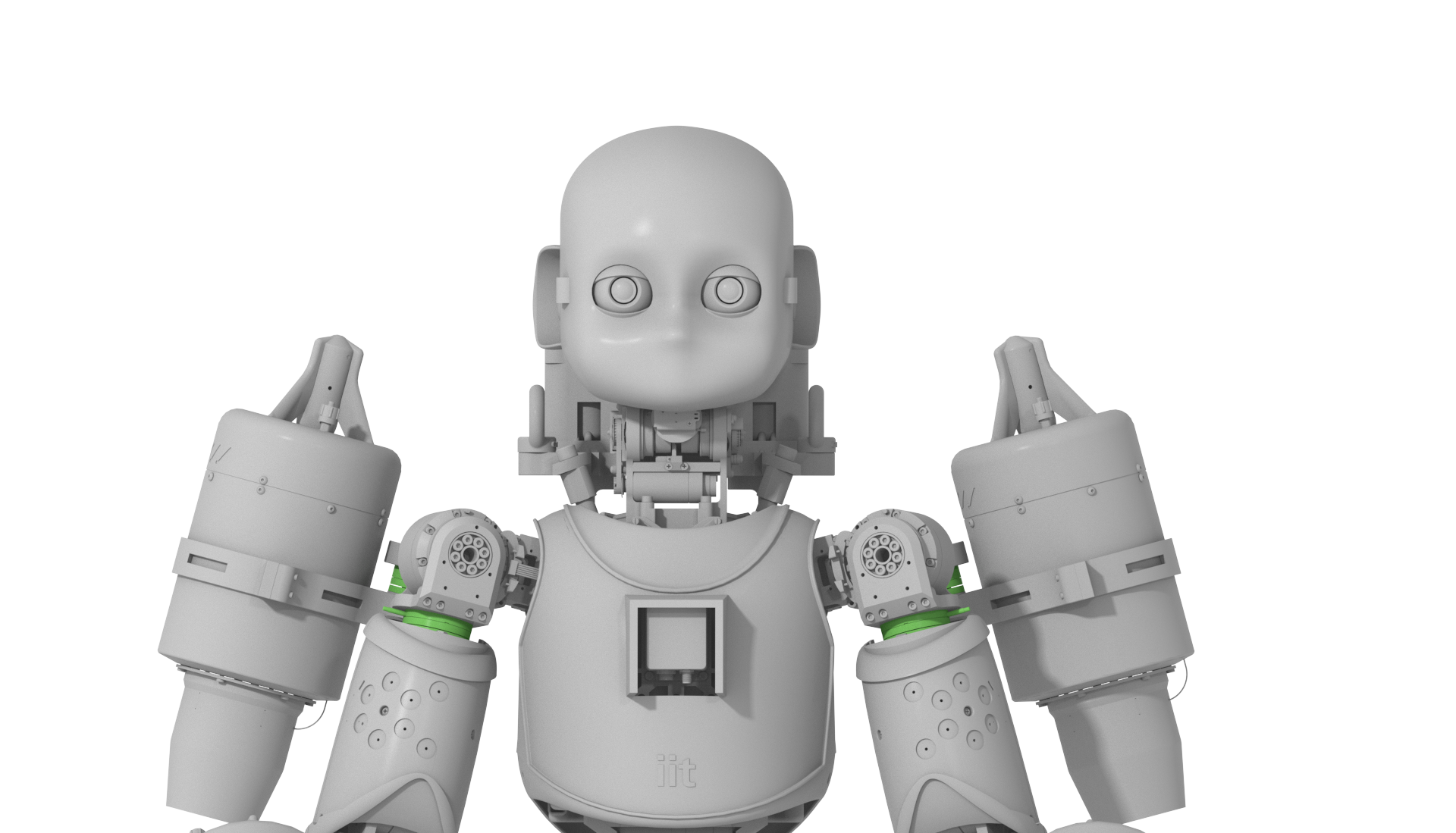}
    \caption{In green: the force-torque sensors mounted on the \textbf{arms} of iRonCub-Mk3.\looseness=-1}
    \label{fig:fts-arms}
\end{figure}

\begin{figure}
    \centering
    \includegraphics[width=0.95\linewidth]{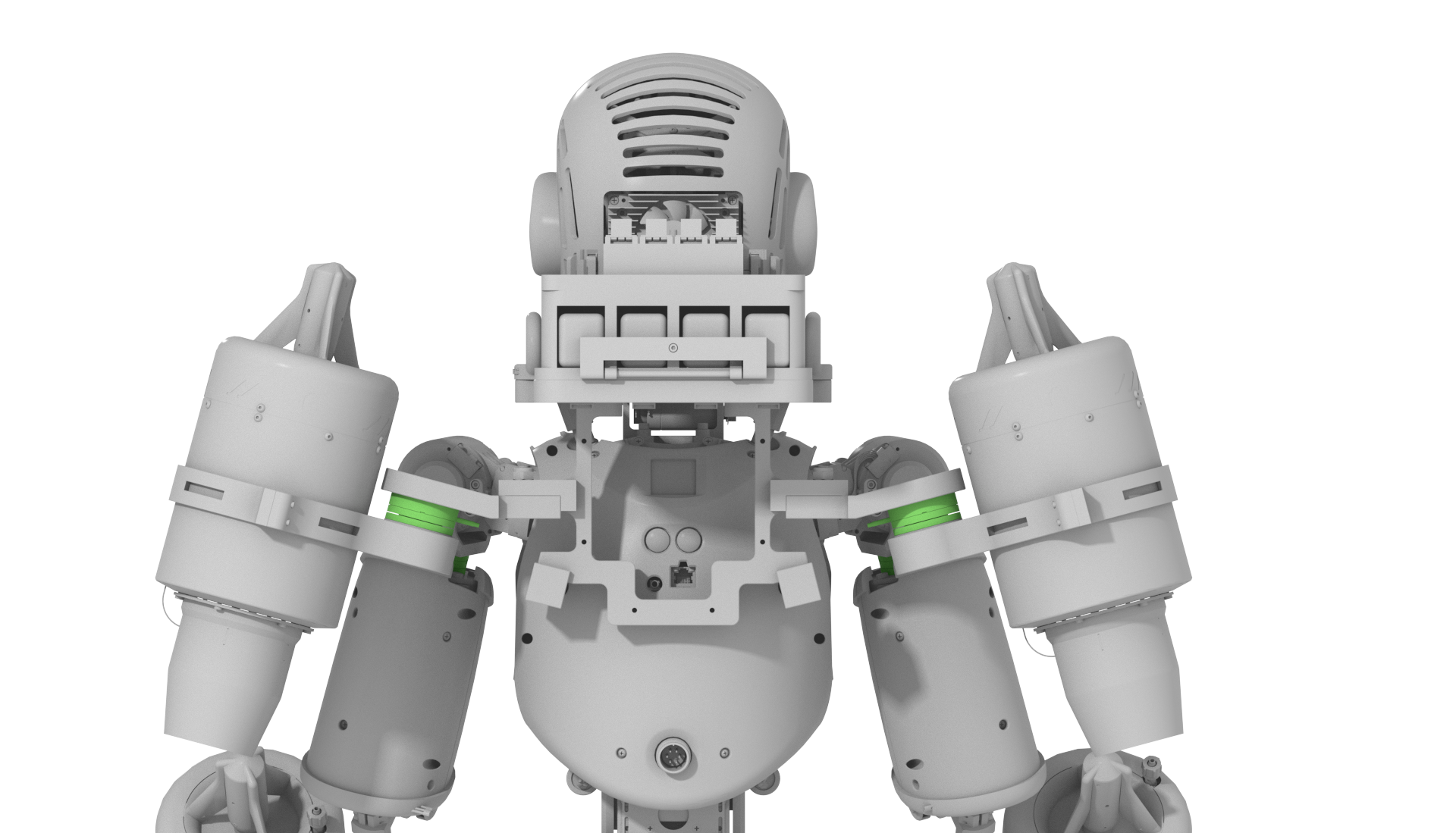}
    \caption{In green: the force-torque sensors mounted on the \textbf{jetpack} of iRonCub-Mk3.\looseness=-1}
    \label{fig:fts-jetpack}
\end{figure}

The \textit{iCub3} robot is already equipped with force-torque sensors on its arms, as illustrated in Fig.~\ref{fig:fts-arms}. In addition, the \textit{iRonCub-Mk3} features two extra force-torque sensors mounted on the jetpack to directly measure the thrust forces generated by the rear jet engines. The mounting configuration of these sensors is shown in Fig.~\ref{fig:fts-jetpack}.

The force-torque sensors' data, combined with the rotation speeds of the jet engines (RPM), provide thrust intensity measurements to the UKF thrust estimation. 

The Jet Control is devoted to sending throttle commands to the turbines, and it operates at 10~Hz, the same frequency at which the low-level controller of the turbines runs.

\subsection{Base Pose Estimation}
\label{subsec:basePoseEstim}

For what concerns the estimation of the base position, we rely on two sensors: one Inertial Measurement Unit (IMU) and a depth camera. We chose the Xsens MTI-670G as IMU and the RealSense8 T265 as depth camera. The Xsens IMU provides the orientation and the angular velocity, while the Realsense measures the position, orientation, linear velocity, and angular velocity.
A UKF combines the measurements coming from the two sensors to provide an estimation of the position, orientation, linear velocity, and angular velocity of the base of the robot at a frequency of $200 Hz$.

\subsection{Flight Controller}
\label{subsec:flightController}

The controller used for the experiment is the Linear Parameter Varying Model Predictive Control (MPC) proposed in \cite{gorbani2025unifiedmultiratemodelpredictive}; this MPC combines the equation of the centroidal momentum \cite{orin2013centroidal} and the nonlinear second-order model proposed in \cite{lerario2020jets} to predict the dynamics of the jet engines, and computes the optimal joints position and jet throttle to track a reference trajectory. In fact, the dynamics of the jets are highly nonlinear, and it has to be taken into account when building the controller. Moreover, this controller is able to deal with the different actuation frequencies of the jets and the joints, computing the control input for the different actuators at a frequency consistent with the actuation bandwidth.

One key limitation of the proposed MPC is its reliance on Euler angles, which are susceptible to singularities. However, in our chosen reference frame, the singularity occurs only when the robot is pitched 90 degrees—i.e., fully horizontal. Since our focus is on Vertical Take-Off and Landing (VTOL), the robot operates far from this condition. Moreover, due to the assumption made in the linearisation, this controller is not suitable for performing aggressive manoeuvres.

\subsubsection{Take-off phase}
\label{subsubsec:takeoff}

During the take-off phase, the robot is in contact with the ground through the feet, so there are two contact forces, one per foot, that should be taken into account when writing the equation of the centroidal momentum \cite{gorbani2025unifiedmultiratemodelpredictive}:
\begin{equation}
    ^{\frameCent} \dot{h} = {A}({q}){T}+ m {g} e_3.
    \label{eq:centroidalMom}
\end{equation}
Since we don't have an adequate estimate of the contact forces, we rewrite \eqref{eq:centroidalMom} in the following way:
\begin{equation}
    ^{\frameCent} \dot{h} = {A}({q}){T}+ \alpha m {g} e_3 ,
\end{equation}
where $\alpha$ is a parameter that regulates the percentage of the body weight that is compensated by the thrust forces. This parameter during the take-off phase starts from $0$ when the turbines are not providing any thrust force, and when it reaches one, the robot detaches from the ground.

\subsection{Simulation environment}
All the software parts are tested first in simulation; the simulations are performed using the Gazebo simulator \cite{koenig2004design}. Thanks to the YARP Gazebo plugins \cite{mingo2014yarp}, we are able to use the same software components both in simulation and on the real robot, since they provide the same interfaces used for the real robot also in Gazebo. 

Simulation-based verification is a critical step, particularly given the safety risks associated with the jet propulsion system, which operates at temperatures around $600^{\circ}$C. Early detection of software faults in simulation mitigates the risk of hardware damage or hazardous conditions during experimental trials. 

\begin{figure}
    \centering
    \includegraphics[width=1.0\linewidth]{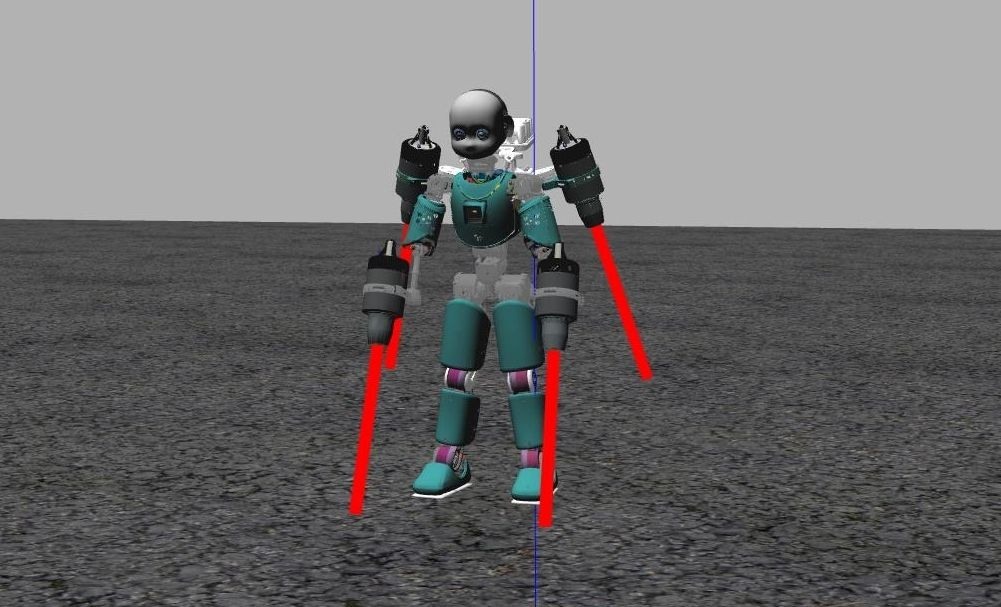}
    \caption{The iRonCub-MK3 robot in the Gazebo simulation environment, where comprehensive testing of its flight control strategies and system interactions is carried out. This virtual setup provides a safe and flexible platform for evaluating performance under various conditions before transitioning to physical experiments.}
    \label{fig:ironcubGazebo}
\end{figure}

This approach also contributes to reducing the \textit{sim-to-real} gap by testing the same software both in simulation and on the real hardware. Moreover, the noise on the measurement induced by the vibration of the turbines is added in simulation; to do that, we computed the standard deviation of the noise from data acquired with the real hardware and added it to the measurements coming from the simulated sensors. Fig.~\ref{fig:ironcubGazebo} shows the iRonCub robot in a simulation in Gazebo while hovering.\looseness=-1

\section{EXPERIMENTAL AREA}
\label{sec:expArea}
\begin{figure}
    \centering
    \includegraphics[width=1.0\linewidth]{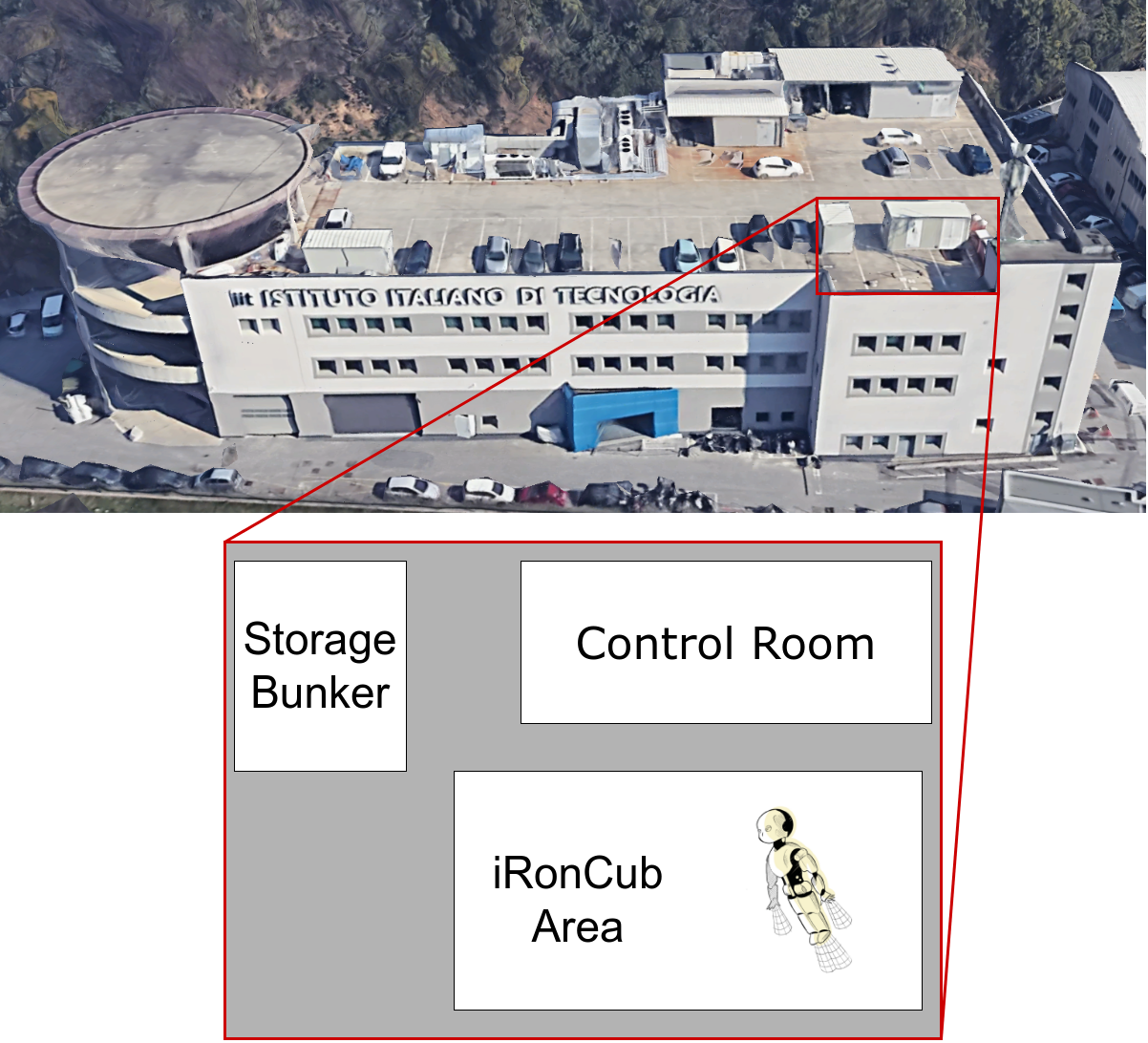}
    \caption{3D view from Google Earth showing the Center for Robotics and Intelligent Systems (CRIS) building at the Italian Institute of Technology, with a zoomed-in focus on the designated flight testing area.}
    \label{fig:cris}
\end{figure}

The experiments are conducted in a flight area on the terrace of the Center for Robotics and Intelligent Systems (CRIS), a key hub of the Italian Institute of Technology. A 3D reconstruction of the building, captured from Google Earth, is shown in Fig.~\ref{fig:cris}.

\begin{figure*}
    \centering
    \includegraphics[trim = 0 0 850 0 , clip, width=0.75\linewidth]{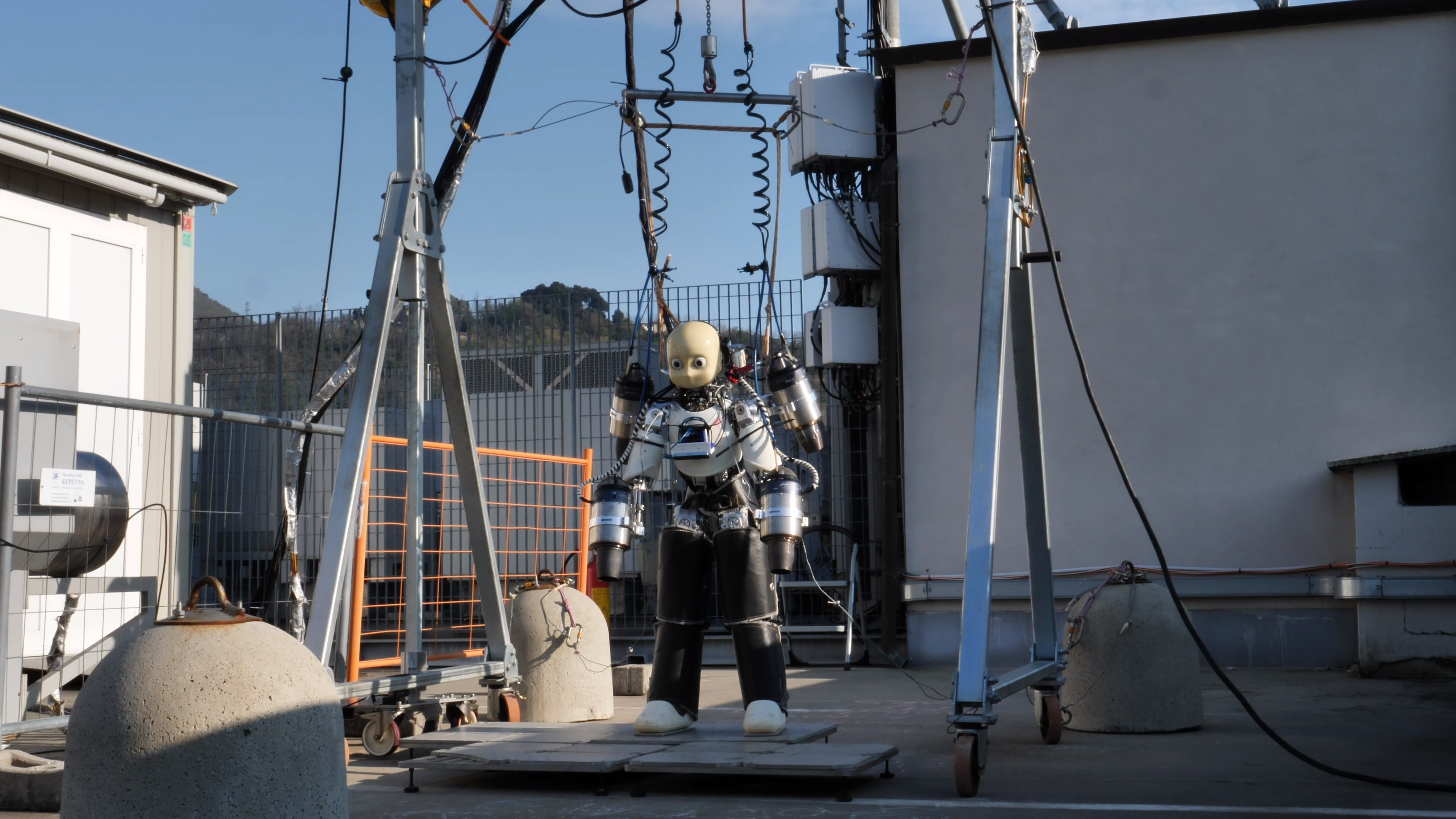}
    \caption{The iRonCub Area includes the iRonCub-Mk3 attached to a two-meter crane. Two concrete bollards behind the robot act as anchor points, with loose metal cables securing the robot to them for safety during flight tests. The lower section of the robot is covered with heat-resistant Aerogel material to shield it from the heat generated by the jet exhaust.}
    \vspace{-0.5cm}
    \label{fig:ironcubarea}
\end{figure*}
The facility consists of three main zones:
\begin{itemize}
    \item iRonCub Area: the primary test zone where experimental flights are executed.
    \item Control Room: The operational hub for software execution and real-time monitoring, located at a safe distance from the flight area.
    \item Storage Bunker: A secure, shielded storage area for equipment and fuel systems necessary for the experiments.
\end{itemize}

Fig.~\ref{fig:ironcubarea} depicts the iRonCub Area, where the experiments are conducted. In this setup, the robot is suspended from a 2-meter-high crane, providing sufficient manoeuvrability while remaining securely tethered to prevent accidents in case of controller failure. To aid heat dissipation and protect the robot from overheating, its lower section is covered with heat-resistant Aerogel material to shield it from the jet exhaust’s high temperatures. Additionally, raised platforms fit with cement tiles are placed on the ground to elevate the robot, minimising heat buildup during the tests. The figure also highlights the fuel line supplying the jet engines, with jet fuel stored in tanks housed within an explosion-proof container.

Inside the Control Room, laptops connect to the robot’s onboard computer via Secure Shell (SSH), allowing remote execution of all control and estimation software modules. A dedicated machine handles data logging and video recording from fixed webcams. The data acquisition system employs the same logging framework described in \cite{dafarra2024icub3}, enhanced to support real-time telemetry visualisation, which aids in monitoring system performance during flight. During each experimental session, at least two team members remain in the Control Room to oversee software operations, while at least one team member stays at a safe distance near the experimental area. This person monitors the experiments directly to compensate for any communication delays from the Control Room and is responsible for reporting any issues that may not be immediately noticeable from inside. \looseness=-1

\begin{figure*}
    \centering
    \includegraphics[width=1.0\linewidth]{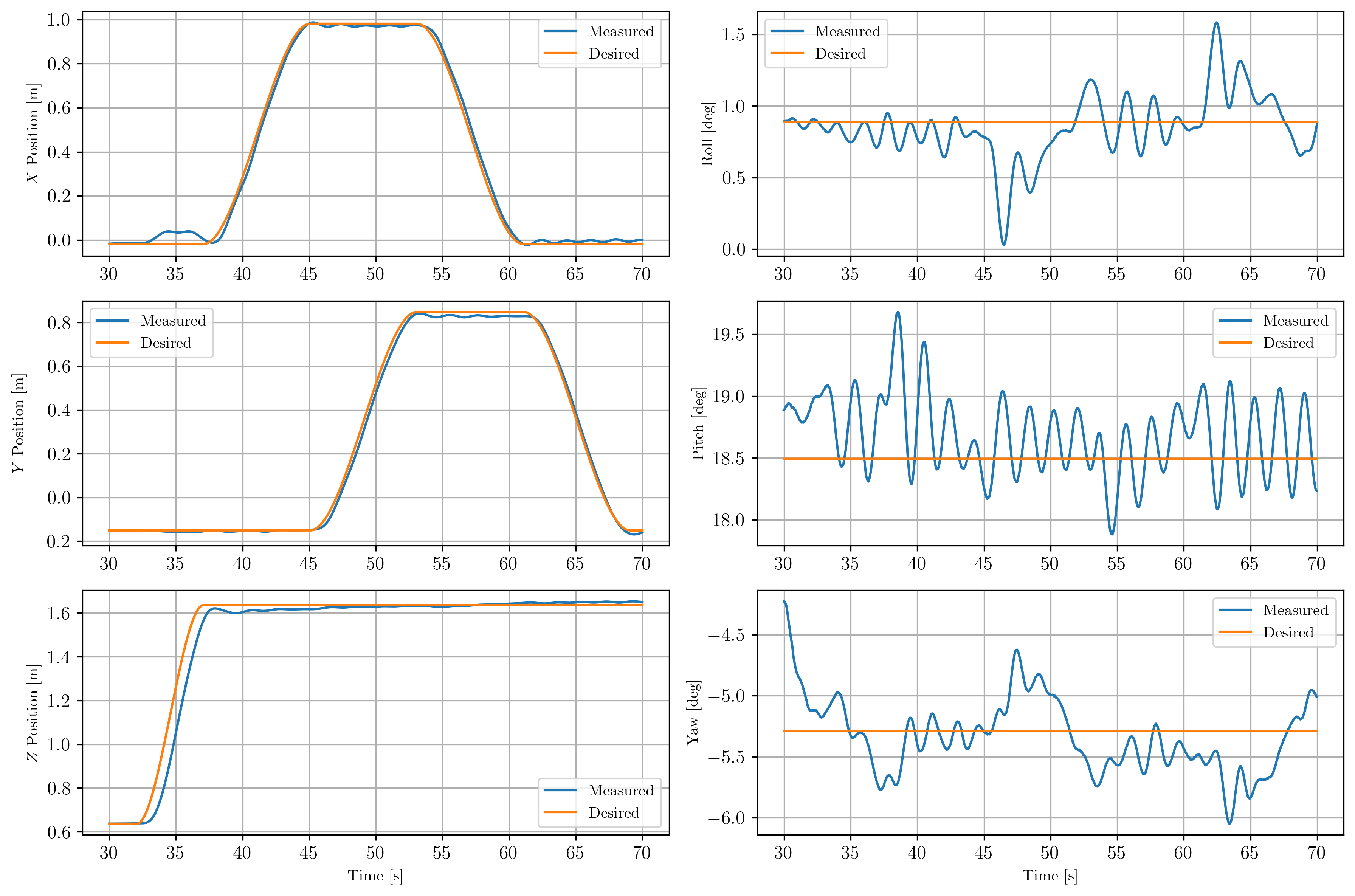}
    \caption{Simulation results depicting the tracking of the centre of mass position and base orientation performing a square trajectory.}
    \label{fig:squareTracking}
\end{figure*}

\begin{figure*}
    \centering
    \includegraphics[width=1.0\linewidth]{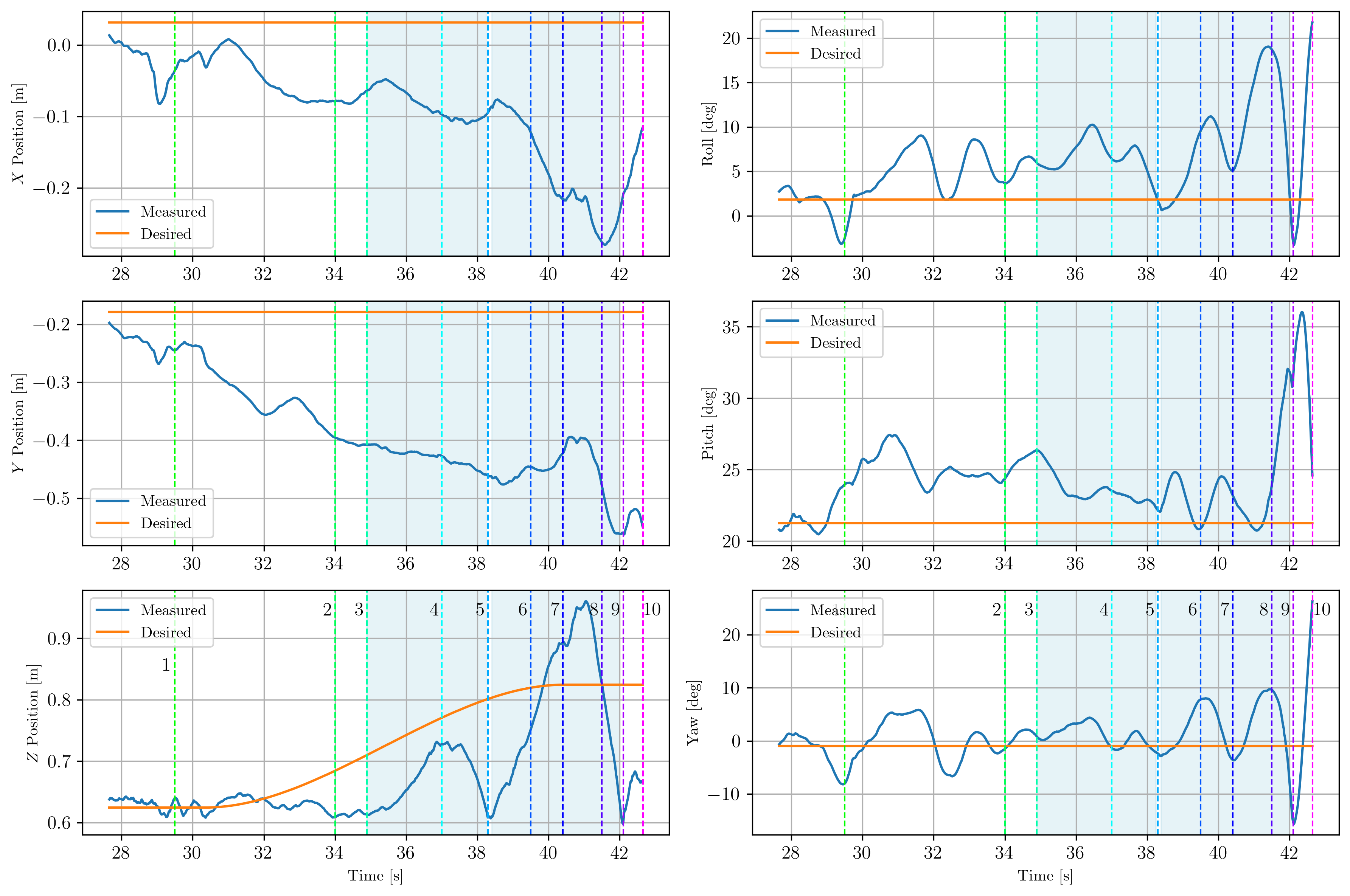}
    \caption{Experimental results illustrating the tracking of the centre of mass position and orientation during take-off with the real robot. The light blue regions indicate the time intervals when the robot was airborne, while the dashed lines represent the frames in Fig. \ref{fig:flight-frames}.}
    \label{fig:tracking}
\end{figure*}
\begin{figure*}
    \centering
    \includegraphics[width=0.95\linewidth]{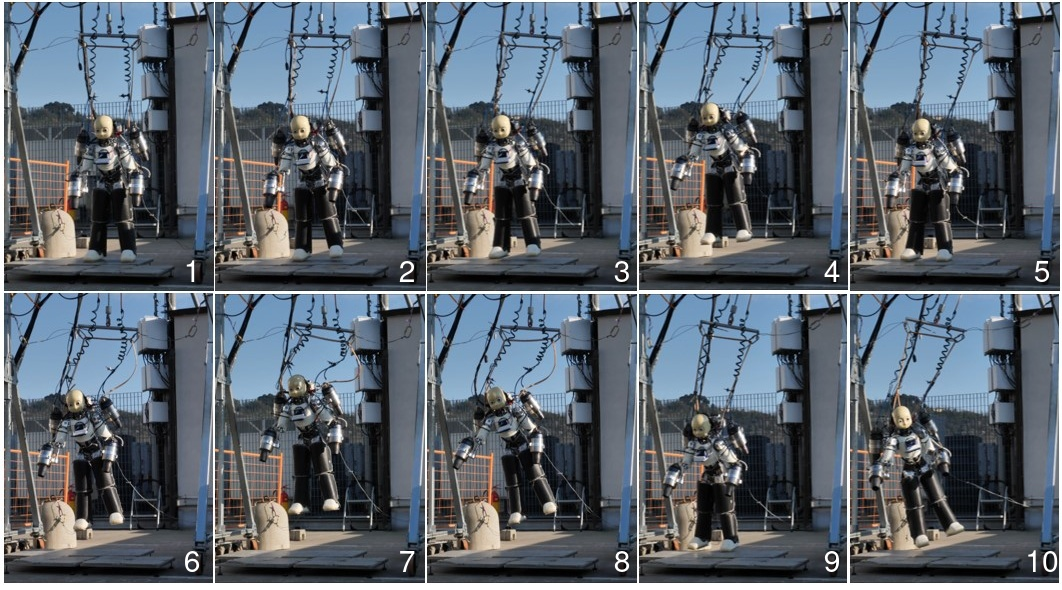}
    \caption{Sequence of frames capturing the real robot performing a take-off. The robot maintains a stable orientation while exhibiting slight horizontal drift. It then begins to ascend, stays airborne for a brief period, momentarily touches down, and subsequently lifts off again. Finally, the robot descends and is automatically shut down by the flight controller due to the orientation error surpassing a predefined threshold.}
    \label{fig:flight-frames}
\end{figure*}
\begin{figure}
    \centering
    \includegraphics[width=1.0\linewidth]{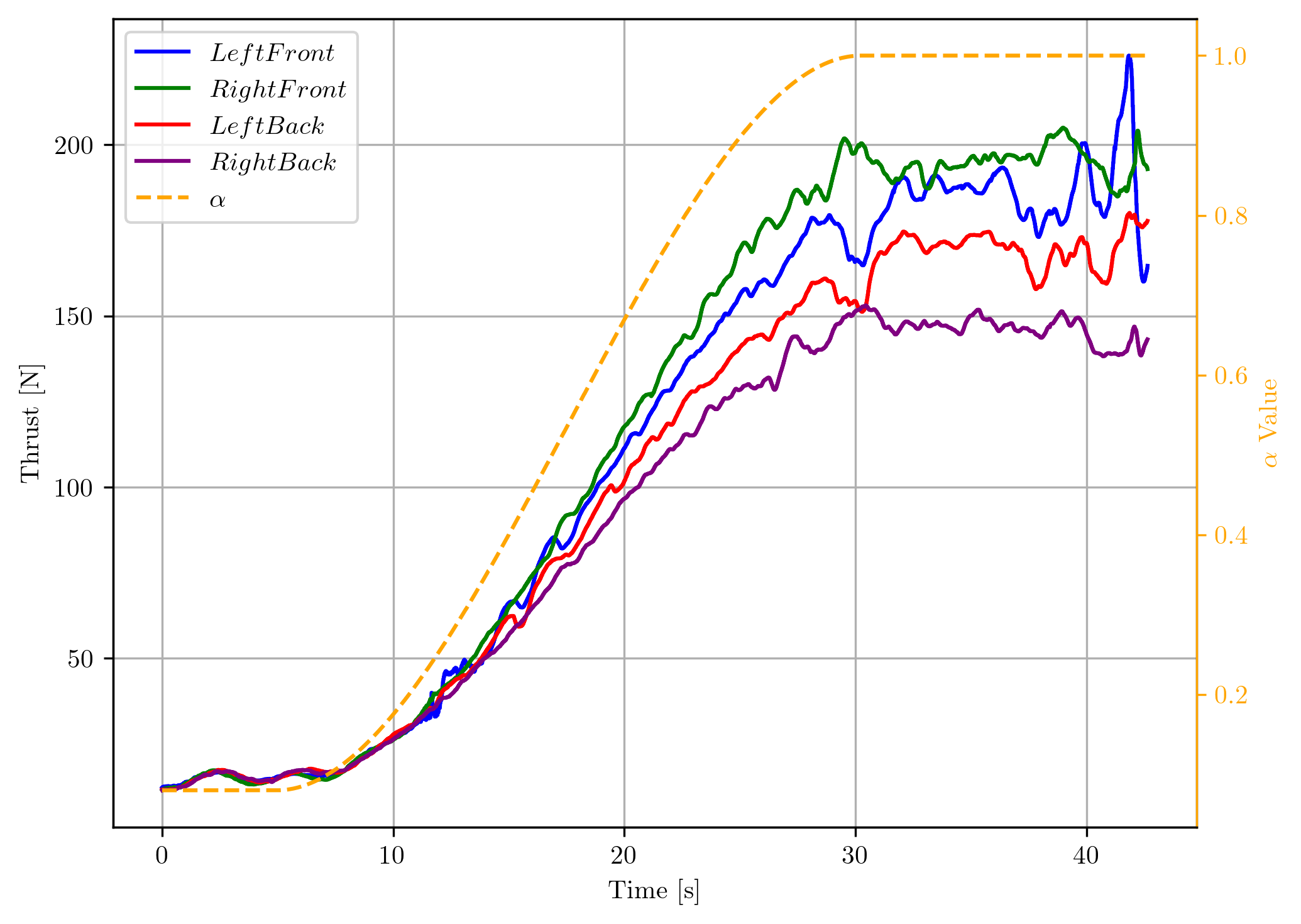}
    \caption{Estimated thrust intensities and in dotted yellow the parameter $\alpha$ that regulates the transition during the take off from the ground to the hovering.}
    \label{fig:thrust}
\end{figure}

\section{RESULTS}
\label{sec:results}
This section presents experimental results obtained both in simulation and on the physical robot. In simulation, we show the capabilities of the controller to track a reference trajectory, while on the real robot, we present the results of a take-off experiment.

\subsection{Simulation results}

First, we show the results that we got in simulations. In order to test the overall software infrastructure, we chose to make the robot perform a take-off, increase its height by about 1 meter and follow a square trajectory; concerning the reference orientation, we chose to keep the desired orientation constant to the initial value. Fig.~\ref{fig:squareTracking} shows the performance of the controller in tracking the trajectory; this is a quite slow and smooth trajectory, and the robot is able to follow it quite well. The Mean Absolute Error (MAE)

\subsection{Experiment on the real robot}

This section presents the experimental results obtained with the physical robot platform. Fig.~\ref{fig:thrust} illustrates the thrust profiles of the four turbines; on the same plot, the value of the parameter $\alpha$ described in \ref{subsubsec:takeoff} is shown. The take-off phase begins approximately at $t = 5$~s and concludes at $t = 30$~s. After this point, the reference trajectory along the vertical ($z$) axis increases, prompting the robot to initiate lift-off from the ground.

Fig.~\ref{fig:tracking} presents the tracking performance of the robot's centre of mass (CoM) and base orientation. Throughout the test, the robot maintains its roll, pitch, and yaw angles within approximately $5^\circ$. Nonetheless, a noticeable drift occurs in the horizontal plane, with a displacement of around $-20~\mathrm{cm}$ along the $x$ axis and approximately $60~\mathrm{cm}$ along the $y$ axis. The $z$ coordinate begins to increase about $2.5~\mathrm{s}$ after the reference command is issued, marking the onset of lift-off. The robot remains airborne for roughly $3~\mathrm{s}$, briefly touches down, and then lifts off again, overshooting the target altitude for another $3~\mathrm{s}$. Finally, the robot loses altitude and is automatically powered down by the flight controller after the orientation error exceeds a specified limit. Fig.~\ref{fig:flight-frames} displays a sequence of frames capturing this flight test. The full video of the experiment can be found at the link below\footnote{\href{https://github.com/ami-iit/paper_gorbani_mohamed_2025_ironcub3}{https://github.com/ami-iit/paper\_gorbani\_mohamed\_2025\_ironcub3}}.

The observed deviations from the expected behaviour are attributed to several factors, primarily related to the sim-to-real gap. Firstly, the model predictive control (MPC) framework relies on a simplified model of the jet dynamics, which fails to accurately capture the highly dynamic behaviour of the turbines. Secondly, the jet propulsion system induces vibrations that propagate through the robot structure and affect the onboard inertial and velocity estimation sensors, leading to noisy measurements that degrade control performance. Furthermore, as discussed in~\cite{mohamed2023NonlinearFTCalib}, the calibration of the force-torque (FT) sensors remains a challenging task, and inaccuracies in thrust estimation further compound the control difficulties.

\section{CONCLUSIONS}
\label{sec:conclusion}

This article provided a status update on the iRonCub project, a jet-powered humanoid robot designed to explore the challenges of aerial humanoid locomotion. The system integrates jet propulsion with a whole-body control architecture to investigate vertical flight using a humanoid structure. We described the design, modelling, and control pipeline of iRonCub-Mk3, and conducted experiments in simulation and with the real robot.

We validated the proposed control and estimation architecture through both simulation and real-world experiments. In simulation, the robot was able to execute a vertical takeoff and follow a predefined square trajectory, demonstrating accurate tracking with low mean absolute errors. These results confirm the effectiveness of the control pipeline in an idealised setting and provide a solid foundation for real-world deployment.

In this paper, we demonstrated a controlled liftoff from the ground, which represents an important step towards full autonomous flight.  Future work will be on repeatable, full take-off and landing in new experimental settings that streamline operations. The presented tests confirmed the viability of the control and estimation framework in a real-world setting and highlighted critical challenges related to system dynamics, estimation, and actuation.

Future work will focus on improving the estimation of the robot's base pose and thrust, both of which are currently subject to uncertainty. Additionally, we aim to enhance the robustness of the Model Predictive Controller (MPC) to better handle modelling inaccuracies and external disturbances. These improvements are essential steps toward achieving stable and autonomous aerial locomotion in humanoid robots.






\balance
\bibliographystyle{unsrt}
\bibliography{paper}






\end{document}